%% file: Template.tex
\documentclass{article}
\usepackage{spconf,amsmath,graphicx,pdfpages}
\usepackage{booktabs}
\usepackage{amsmath,amssymb,amsfonts}
\usepackage{algorithmic}
\usepackage{graphicx}
\usepackage{textcomp}
\usepackage{xcolor}
\usepackage{multirow}
\usepackage[normalem]{ulem}
\usepackage{colortbl}
\usepackage{float}
\title{CLIP-AE: CLIP-assisted Cross-view Audio-Visual Enhancement for Unsupervised Temporal Action Localization}
\name{
Rui Xia$^{1\star}$, 
Dan Jiang$^{2\star}$, 
Quan Zhang$^{1}$, 
Ke Zhang$^{1\dagger}$, 
Chun Yuan$^{1\dagger}$
\thanks{
\hangindent=2em
\hangafter=1
$\star$These authors contributed equally to this work.\\
$\dagger$Corresponding Authors, E-mail: ke-zhang19@mails.tsinghua.edu.cn, yuanc@sz.tsinghua.edu.cn\\
}
}

\address{
$^{1}$ Shenzhen International Graduate School, Tsinghua University, Beijing, China\\
$^{2}$ School of Vehicle and Mobility, Tsinghua University, Beijing, China
} 
\begin{document}

%
\maketitle
\begin{abstract}
Temporal Action Localization (TAL) has garnered significant attention in information retrieval. Existing supervised or weakly supervised methods heavily rely on labeled temporal boundaries and action categories, which are labor-intensive and time-consuming. Consequently, unsupervised temporal action localization (UTAL) has gained popularity. However, current methods face two main challenges: 1) Classification pre-trained features overly focus on highly discriminative regions; 2) Solely relying on visual modality information makes it difficult to determine contextual boundaries. To address these issues, we propose a CLIP-assisted cross-view audiovisual enhanced UTAL method. Specifically, we introduce visual language pre-training (VLP) and classification pre-training-based collaborative enhancement to avoid excessive focus on highly discriminative regions; we also incorporate audio perception to provide richer contextual boundary information. Finally, we introduce a self-supervised cross-view learning paradigm to achieve multi-view perceptual enhancement without additional annotations. Extensive experiments on two public datasets demonstrate our model's superiority over several state-of-the-art competitors.
\end{abstract}
\section{Introduction}

With the rapid growth of social media videos, video retrieval has become a popular topic in information retrieval. Traditional methods primarily focus on identifying the most relevant videos based on a given query \cite{zhai2020two}. However, users are often more interested in specific actions within video clips, making temporal action localization (TAL) in untrimmed videos essential \cite{yun2024weakly}.

\begin{figure}[!t]
\centering
\includegraphics[width=1.0\linewidth,height=4cm]{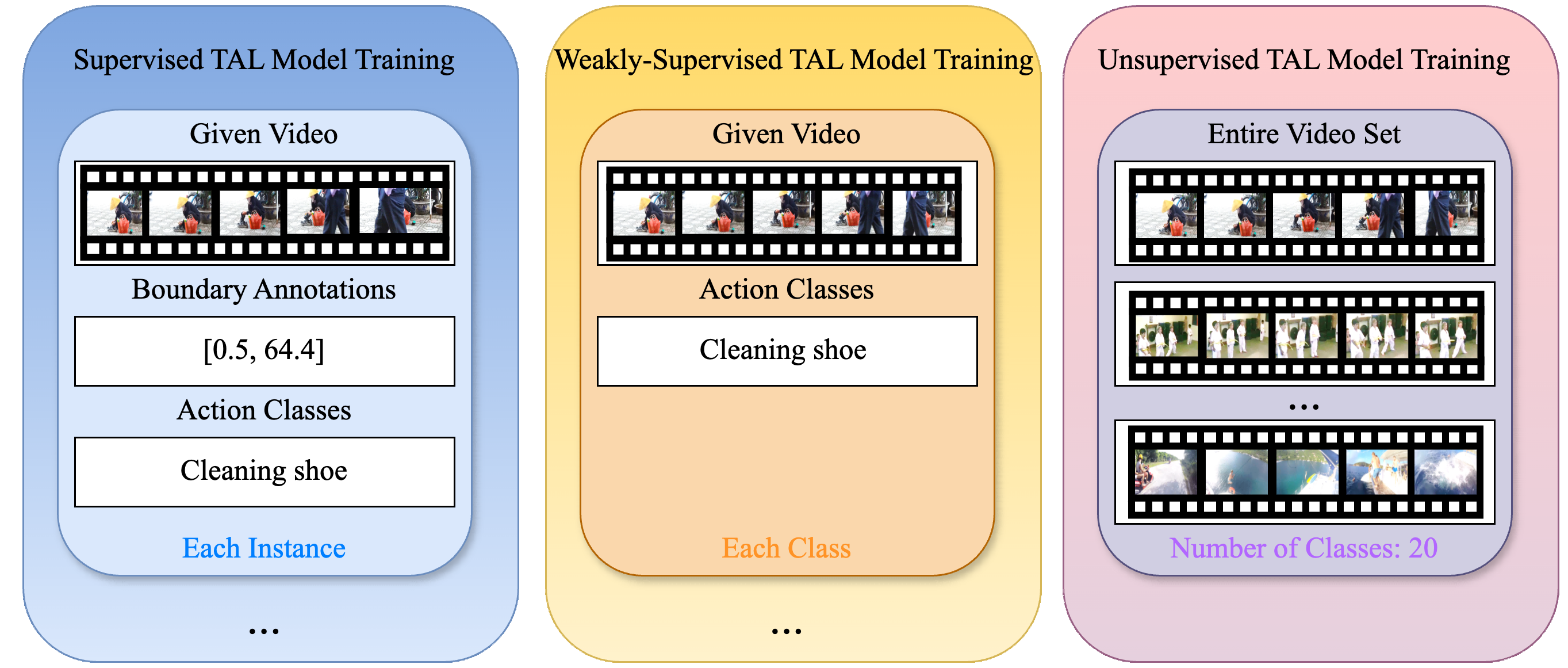}
\caption{Temporal action localization under various levels of supervision}
\label{fig:1}
\end{figure}

Fully supervised TAL methods, as illustrated in Figure \ref{fig:1}, involve time-consuming and error-prone temporal boundary annotations, and subjective labeling can negatively impact localization performance. While weakly supervised approaches reduce the cost of boundary annotations \cite{wang2023two,zhang2024can,Zhang2025Rethinking}, they still require action category labels, which are labor-intensive.

To overcome these challenges, recent research has shifted towards unsupervised TAL (UTAL) \cite{liu2023apsl}, which relies only on the number of action categories within the video dataset for training. Gong et al. and Yang et al. introduced the Temporal Class Activation Map (TCAM) \cite{gong2020learning} and Uncertainty-guided Collaborative Training (UGCT) \cite{yang2022uncertainty} models, respectively, adopting a two-stage iterative process of clustering and localization. These methods generate video-level pseudo-labels followed by model training. Compared to fully or weakly supervised TAL, which require extensive labeling, UTAL offers greater scalability to meet the growing demands of video understanding

\begin{figure*}[t]
\centering
\includegraphics[width=1.0\linewidth]{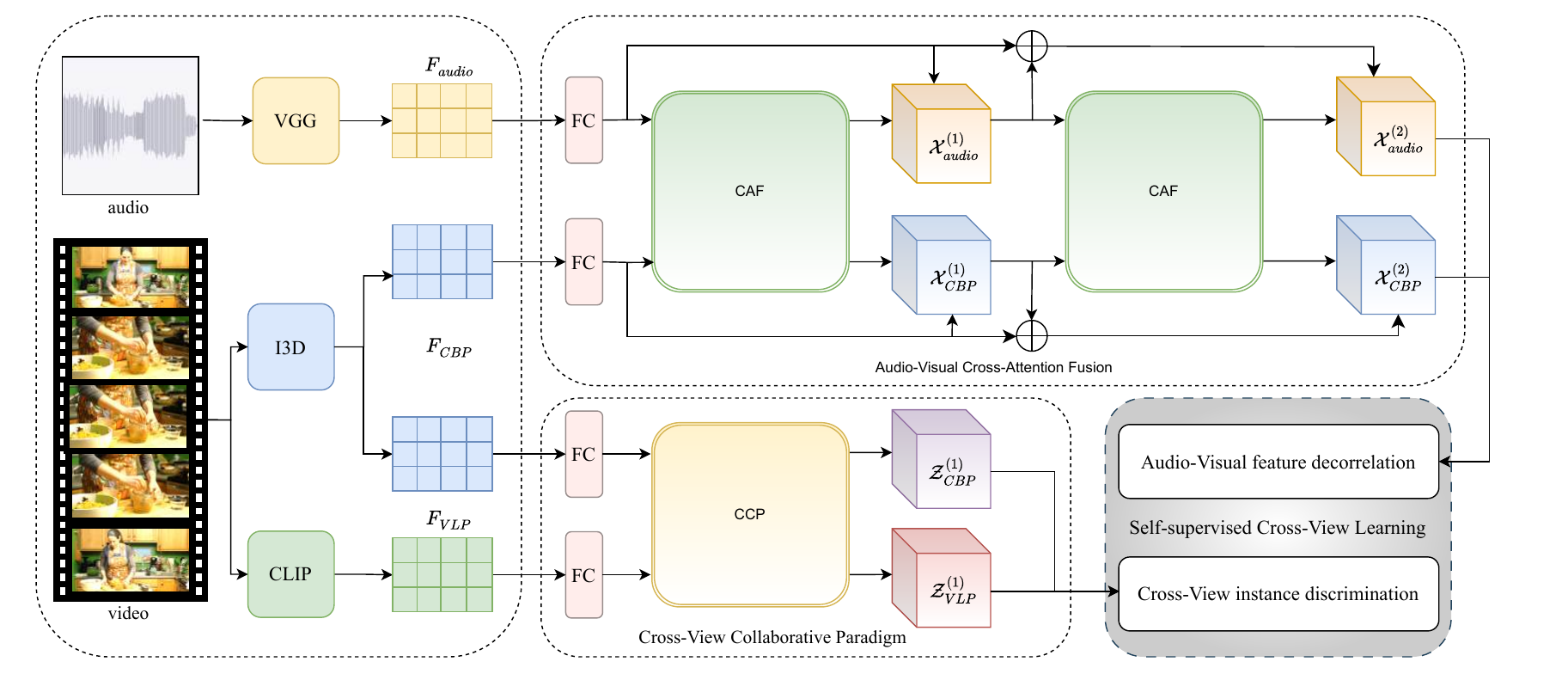}
\caption{CLIP-assisted cross-view audio-visual perception enhancement learning method}
\label{fig:pipeline}
\end{figure*}


Despite the significance of UTAL, it faces two challenges: 1) Classification pre-trained features overly focus on highly discriminative regions. The features used in UTAL for clustering and localization are extracted based on classification pre-training, which often only needs to focus on key frames of action instances for classification tasks, lacking sufficient attention to the complete action interval. 2) Solely relying on visual modality information makes it difficult to determine contextual boundaries. Previous work relied solely on visual features for UTAL, completely ignoring audio information in videos. When encountering actions strongly associated with audio (e.g., blowing a horn, singing), audio is crucial for determining contextual boundaries.

To address these challenges, we propose a novel CLIP-assisted cross-view audiovisual perception enhancement model, named CLIP-AE. Inspired by VLP features that often focus on as complete a temporal action interval as possible, which compensates for the incompleteness brought by CBP features, we propose a cross-view collaboration paradigm (CCP) of CBP and VLP; inspired by audiovisual perception localization methods that use audio information to assist video localization, we propose a cross-attention fusion paradigm for audiovisual features, adopting a multi-stage dense connection approach to achieve full fusion of audiovisual features. Finally, we introduce a self-supervised cross-view learning paradigm, leveraging instance discrimination and feature decorrelation as pretext tasks to achieve multi-view perceptual enhancement without additional annotations.

The main contributions of our proposed CLIP-AE can be summarized as follows:
\begin{itemize}
\item We propose a CLIP-assisted cross-view audiovisual perception enhancement learning method for UTAL, introducing CLIP and audio information into the UTAL task for the first time.
\item We propose a cross-attention fusion module for audiovisual features and a cross-view collaboration paradigm of CBP and VLP, addressing the localization inaccuracy caused by previous work that only used classification pre-trained visual features.
\item We introduce a self-supervised cross-view learning paradigm to achieve multi-view perceptual enhancement without additional annotations.
\item We conducted extensive comparative experiments, ablation studies, and visualizations to verify the excellent performance of our model.
\end{itemize}

\section{Method}

Our CLIP-assisted cross-view audio-visual enhancement framework is shown in Figure \ref{fig:pipeline}. Pre-trained audio, VLP, and CBP feature extractors first extract $F_{audio}$, $F_{CBP}$, and $F_{VLP}$ features. These are processed through the Audio-Visual Cross Attention Fusion module to generate fused audio-visual features. Simultaneously, $F_{CBP}$ and $F_{VLP}$ are input into the Cross-View Collaboration Paradigm to produce enhanced visual features. The outputs from both modules are then passed to the self-supervised cross-view learning module, enabling representation learning through two proxy tasks.

\subsection{Audio-Visual Cross-Attention Fusion}

Fusing multiple modalities can provide more comprehensive information, but this fusion may lead to the loss of modality-specific details. To effectively integrate audio and visual features, we designed an audio-visual cross-attention fusion module based on a multi-stage dense cross-attention mechanism. This mechanism allows each modality to independently learn features that encode inter-modality correlations while retaining intra-modality information.

As shown in the figure \ref{fig:pipeline}, we first encode the audio feature $F_{audio}$ and visual features $F_{CBP}$ through fully connected layers specific to each modality into $X_{audio}=(x^{l}_{audio})^{L}_{l=1}$ and $X_{CBP}=(x^{l}_{CBP})^{L}_{l=1}$, where $x^{l}_{audio}$ and $x^{l}_{CBP}$ are elements of $R^{d_{x}}$. We then compute the cross-correlation matrix of $X_{audio}$ and $X_{CBP}$ to assess the inter-modality correlation. To bridge the heterogeneity gap between the two modalities, we introduce a learnable weight matrix and calculate the cross-correlation as follows: 
\begin{equation}
\Lambda=X_{\text {audio }}^{T} W X_{\mathrm{CBP}}
\end{equation}
where $\Lambda \in \mathbb{R}^{L\times L}$, $x_{\text{audio}}^{l}$ and $x_{\mathrm{CBP}}^{l}$ are L2 normalized prior to cross-correlation computation.

In the cross-correlation matrix, a higher coefficient indicates a stronger correlation between audio and visual features. Consequently, the l-th column of $\Lambda$ corresponds to the correlation between $x^{l}_{CBP}$ and L audio segment features. Using this, we perform a softmax operation on the columns of $\Lambda$ and $\Lambda^{T}$ to generate cross-attention weights $A_{\text {audio }}$ and $A_{\mathrm{CBP}}$. For each modality, these attention weights are applied to re-weight the segment features, enhancing their discriminability in the context of the other modality. Specifically, the attention-weighted features and are derived using the following formula: 
\begin{equation}
\tilde{X}_{\text {audio }}=X_{\text {audio }} A_{\text {audio }} \quad \text { and } \quad \tilde{X}_{\mathrm{CBP}}=X_{\mathrm{CBP}} A_{\mathrm{CBP}}
\end{equation}

It is crucial to note that each modality guides the other through attention weights, ensuring that meaningful intra-modality information is preserved during cross-attention. To deeply explore cross-modal information, we apply cross-attention iteratively. However, to prevent the suppression of original modality-specific features during the multi-stage cross-attention process, we incorporate dense skip connections. At the t-th stage, we implement dense skip connections as follows: 
\begin{equation}
\mathcal{X}_{\text {audio }}^{(t)}=\tanh \left(\sum_{i=0}^{t-1} \mathcal{X}_{\text {audio }}^{(i)}+\tilde{X}_{\text {audio }}^{(t)}\right)
\end{equation}
where ${ }^{\mathcal{X}_{\text {audio }}^{(0)}}$ represents $X_{\text {audio }}$, and tanh denotes the hyperbolic tangent activation function. The same operation applied to $\mathcal{X}_{\text {audio }}^{(t)}$ is used to generate the perceptual features of the visual modality $\mathcal{X}_{\mathrm{CBP}}^{(t)}$. 


\subsection{Cross-View Collaborative Paradigm}
The VLP view's features are designed to capture complete temporal action intervals as thoroughly as possible. By establishing collaborative learning between VLP and CBP views, we can ameliorate the incomplete localization issue caused by classification-based pre-training. The cross-view collaboration between these two views involves the following steps:  
First, we sequentially utilize extractors based on classification pre-training and visual language pre-training to obtain the features of the CBP and VLP views, denoted as $X_{\mathrm{CBP}}=\left\{\boldsymbol{x}_{\mathrm{CBP}}\right\}_{t=1}^{T}$, $X_{\mathrm{VLP}}=\left\{\boldsymbol{x}_{\mathrm{VLP}}\right\}_{t=1}^{T}$. We then concatenate the two into M:
\begin{equation}
\boldsymbol{M}=\left[X_{\mathrm{CBP}} ; X_{\mathrm{VLP}}\right]
\end{equation}

The cross-view collaborative CBP view features are computed through a multi-modal attention mechanism:
\begin{equation}
\begin{gathered}
Z_{\mathrm{CBP}}=\operatorname{Softmax}\left(\frac{\boldsymbol{Q}_{\mathrm{CBP}} \boldsymbol{K}_{\mathrm{CBP}}^{T}}{\sqrt{d}}\right) \boldsymbol{V}_{\mathrm{CBP}} \\
Q_{\text{CBP}} = M W^{Q}, \quad K_{\text{CBP}} = M W^{K}, \quad V_{\text{CBP}} = X_{\text{CBP}} W^{V}
\end{gathered}
\end{equation}

Similarly, we can derive the cross-view collaborative VLP view features:
\begin{equation}
Z_{\mathrm{VLP}}=\operatorname{Softmax}\left(\frac{\boldsymbol{Q}_{\mathrm{VLP}} \boldsymbol{K}_{\mathrm{VLP}}^{T}}{\sqrt{d}}\right) \boldsymbol{V}_{\mathrm{VLP}}
\end{equation}
\subsection{Self-Supervised Cross-View Learning}
\textbf{Audio-visual feature decorrelation:} The representation learning of audio-visual cross-attention fusion is achieved through feature decorrelation, aiming to orthogonalize each dimension of the feature representation. The decorrelation process is as follows:
\begin{small}
\begin{equation}
\mathcal{L}_{\text {de-cor}}=\left\|\mathcal{X}_{\text {audio}}\left(\mathcal{X}_{\text {audio}}\right)^{T}-I\right\|^{2}+\left\|\mathcal{X}_{\mathrm{CBP}}\left(\mathcal{X}_{\mathrm{CBP}}\right)^{T}-I\right\|^{2}
\end{equation}
\end{small}
\textbf{Cross-view instance discrimination:} The representation learning of the cross-view collaboration paradigm is anchored in instance discrimination. For the cross-view collaborative features $Z_{\mathrm{VLP}}$ and $Z_{\mathrm{CBP}}$, we first perform average pooling to obtain $z_{\mathrm{VLP}}, z_{\mathrm{CBP}}$, and then proceed with the instance discrimination process: 
\begin{equation}
\begin{aligned}
\mathcal{L}_{\text {ins-dis }}= & -\log \frac{\exp \left(z_{\mathrm{VLP}}^{+} \cdot m_{\mathrm{VLP}}^{+} / \tau\right)}{\sum_{i=1}^{N} \exp \left(x_{\mathrm{VLP}}^{+} \cdot m_{\mathrm{VLP}}^{i} / \tau\right)} \\
& -\log \frac{\exp \left(x_{\mathrm{CBP}}^{+} \cdot m_{\mathrm{CBP}}^{+} / \tau\right)}{\sum_{i=1}^{N} \exp \left(x_{\mathrm{CBP}}^{+} \cdot m_{\mathrm{CBP}}^{i} / \tau\right)}
\end{aligned}
\end{equation}
where $\tau$ is the temperature hyperparameter (set to 1.0), $z_{\mathrm{VLP}}^{+}$,$m_{\mathrm{VLP}^{+}}$ and $z_{\mathrm{CBP}}^{+}$, $m_{\mathrm{CBP}}^{+}$ are the positive pairs of CBP and VLP views, and $\left\{m_{i}\right\}_{i=1}^{N}$ is the memory bank that maintains the momentum mechanism for each view. 

The final self-supervised loss function is defined as:
\begin{equation}
    \mathcal{L}_{\text {self }}=\mathcal{L}_{\text {de-cor }}+\mathcal{L}_{\text {ins-dis }}
\end{equation}

\section{Experiments}

\input{table/table01}
\input{table/table02}

\input{table/table03}

\subsection{Datasets}




\textbf{THUMOS14} consists of 200 validation and 213 test videos across 20 action classes, averaging 15 action segments per video. Validation data is used for training, and test data for evaluation. \textbf{ActivityNet v1.2} includes 4819 training, 2383 validation, and 2480 test videos. The validation set is used as test data, with an average of 1.5 action segments per video across 100 categories.

\subsection{Implementation Details}
Inspired by the widespread application of deep learning across various fields\cite{zhang2025imdprompter,chen2025gim,qi2025seeing,Zhang_2024_CVPR}, we employed a variety of deep learning-based feature extractors. For CBP features, we utilized a pre-trained I3D network; for VLP features, we employed CLIP based on VIT-B; and for audio features, we used a VGG-like network pre-trained on AudioSet. Unless otherwise stated, the UTAL baseline is UGCT.
\subsection{Main Results}


We compared our method with several state-of-the-art UTAL and WTAL approaches. As shown in Table 1, our method achieves an average mAP (0.1:0.7) of 41.7\% on the THUMOS14 dataset, outperforming UGCT by 1.5\% and APSL by 6.5\%. Table 2 shows the localization performance on ActivityNet v1.2, with the 'AVG' column representing the average mAP across IoU intervals from 0.5 to 0.95. In the UTAL setting, our CLIP-AE achieves an average mAP of 28.9\%, surpassing all unsupervised methods and remaining competitive with recent weakly supervised approaches.

\subsection{Ablation Studies}

We conducted ablation experiments to assess the contribution of each module, as shown in Table 3. Without any proposed components, the average mAP is 22.7\%. Adding the CAF module, which integrates audio information, increases the mAP to 25.1\%, a 2.4\% improvement, highlighting the effectiveness of CAF in constructing audio-visual features. Incorporating the CCP module, which integrates VLP view information, further raises the mAP to 26.5\%, a 3.8\% increase, indicating that VLP view compensates for CBP view's limitations. With both CAF and CCP, the mAP reaches 28.9\%, a 6.2\% improvement, demonstrating their synergistic effect in cross-view audio-visual enhancement.

\section{Conclusion}
We propose CLIP-AE, a CLIP-assisted cross-view audiovisual perception learning paradigm, to address unsupervised temporal action localization (UTAL). This is the first time CLIP and audio are incorporated into UTAL. Our approach introduces a cross-view collaboration between CBP and VLP, enabling mutual enhancement. We also present a cross-attention fusion module for full audiovisual feature integration. Additionally, a self-supervised cross-view learning strategy is introduced for multi-view perceptual enhancement without extra annotations. Extensive experiments, ablation studies, and visualizations confirm the effectiveness of our method.

\newpage

\section{Acknowledgement}
This work was supported by the National Key R\&D Program of China (2022YFB4701400/4701402), SSTIC Grant(KJZD2
0230923115106012,KJZD20230923114916032,GJHZ20240
218113604008), Beijing Key Lab of Networked Multimedia and National Natural Science Foundation of China under Grant 62202302.

\bibliographystyle{IEEEbib}
\bibliography{refs}

\end{document}

%% file: table/table01.tex
\begin{table*}[h]
\centering
\caption{Performance comparison with SOTA methods on THUMOS’14. * denotes the re-implementation for UTAL in UGCT}
\label{tab:my-table-1}
\resizebox{0.9\textwidth}{!}{
\begin{tabular}{c|c|c|ccccccc|ccc}
\hline
                               &                          &                                               & \multicolumn{7}{c|}{mAP@IoU(\%)}               & \multicolumn{3}{c}{AVG}    \\ \cline{4-13} 
\multirow{-2}{*}{Supervision}  & \multirow{-2}{*}{Method} & \multirow{-2}{*}{{Pub.}} & 0.1  & 0.2  & 0.3  & 0.4  & 0.5  & 0.6  & 0.7  & 0.1:0.5 & 0.3:0.7 & 0.1:0.7 \\ \hline
                               & TCAM\cite{gong2020learning}                 & CVPR2020                                      & -    & -    & 46.9 & 38.9 & 30.1 & 19.8 & 10.4 & -       & 29.2    & -       \\
                               & UGCT\cite{yang2022uncertainty}                     & TPAMI2022                                     & 70.3 & 65.3 & 57.9 & 47.8 & 35.8 & 23.3 & 11.1 & 55.4    & 35.2    & 44.5    \\
                               & ARSL\cite{liu2023apsl}                     & INS2023                                       & 69.1 & 62.4 & 53.7 & 43.6 & 33.6 & 23.8 & 12.8 & 52.5    & 33.5    & 42.7    \\
                               & CASE\cite{liu2023revisiting}                     & ICCV2023                                      & 72.3 & -    & 59.2 & -    & 37.7 & -    & 13.7 & 56.4    & 36.9    & 45.7    \\
                               & Wang et al.\cite{wang2023two}              & CVPR2023                                      & 73.0 & 68.2 & 60.0 & 47.9 & 37.1 & 24.4 & 12.7 & 57.2    & 36.4    & 46.2    \\
                               & ISSF\cite{yun2024weakly}                     & AAAI2024                                      & 72.4 & 66.9 & 58.4 & 49.7 & 41.8 & 25.5 & 12.8 & 57.8    & 37.6    & 46.8    \\
\multirow{-7}{*}{Weakly}      & CLIP-AE                  & -                                             & 74.5 & 69.5 & 60.3 & 51.4 & 38.1 & 26.7 & 15.4 & 58.8    & 38.4    & 48.0    \\ \hline
                               & STPN*\cite{nguyen2018weakly}                     & CVPR2018                                      & 50.1 & 45.8 & 40.6 & 32.3 & 20.9 & 10.7 & 4.6  & 37.9    & 21.8    & 29.3    \\
                               & WSAL-BM*\cite{nguyen2019weakly}                  & CVPR2019                                      & 57.7 & 52.4 & 46.4 & 37.1 & 26.1 & 16.0 & 6.7  & 43.9    & 26.5    & 34.6    \\
                               & TSCN*\cite{zhai2020two}                     & CVPR2020                                      & 57.1 & 51.6 & 43.9 & 35.3 & 26.0 & 15.7 & 6.0  & 42.8    & 25.4    & 33.7    \\
                               & TCAM\cite{gong2020learning}                       & CVPR2020                                      & -    & -    & 39.6 & 32.9 & 25.0 & 16.7 & 8.9  & -       & 24.6    & -       \\
                               & UGCT\cite{yang2022uncertainty}                     & TPAMI2022                                     & 63.4 & 57.8 & 51.7 & 44.0 & 32.8 & 21.6 & 10.1 & 49.9    & 32.0    & 40.2    \\
                               & APSL\cite{liu2023apsl}                     & INS2023                                       & 57.7 & 52.4 & 44.1 & 35.9 & 27.9 & 18.5 & 10.0 & 43.6    & 27.3    & 35.2    \\
\multirow{-7}{*}{Unsupervised} & CLIP-AE                  & -                                             & 65.1 & 60.4 & 52.7 & 44.5 & 33.2 & 22.3 & 13.6 & 51.2    & 33.3    & 41.7  \\ \hline
\end{tabular}
}
\end{table*}

%% file: table/table02.tex
\begin{table}[h]
\centering
\caption{ Performance comparison with SOTA methods on
ActivityNet v1.2. * denotes the re-implementation for UTAL in UGCT.}
\label{tab:my-table-2}
\resizebox{0.45\textwidth}{!}{
\begin{tabular}{c|c|ccc|c}
\hline
\multirow{2}{*}{\textbf{Supervision}} &
  \multirow{2}{*}{\textbf{Method}} &
  \multicolumn{3}{c|}{\textbf{mAP@IoU(\%)}} &
  \multirow{2}{*}{\textbf{AVG}} \\
 &
   & 0.5 & 0.75 & 0.95 &  \\
\hline
\multirow{5}{*}{Weakly} & TCAM\cite{gong2020learning}      & 40.0 & 25.0 & 4.6  & 24.6 \\
                         & UGCT\cite{yang2022uncertainty}    & 43.1 & 26.6 & 6.1  & 26.9 \\
                         & APSL\cite{liu2023apsl}    & 44.3 & 28.5 & 6.2  & 28.2 \\
                         & PMIL\cite{ren2023proposal}    & 44.2 & 26.1 & 5.3  & 26.5 \\
                         & CLIP-AE & 50.1 & 30.3 & 6.8  & 30.5 \\
\hline
\multirow{7}{*}{Unsupervised} & STPN\cite{nguyen2018weakly}    & 28.2 & 16.5 & 3.7  & 16.9 \\
                              & WSAL-BM\cite{nguyen2019weakly} & 28.5 & 17.6 & 4.1  & 17.6 \\
                              & TSCN\cite{zhai2020two}    & 22.3 & 13.6 & 2.1  & 13.6 \\
                              & TCAM\cite{gong2020learning}    & 35.2 & 21.4 & 3.1  & 21.1 \\
                              & UGCT\cite{yang2022uncertainty}    & 37.4 & 23.8 & 4.9  & 22.7 \\
                              & APSL\cite{liu2023apsl}    & 43.7 & 28.1 & 5.8  & 27.6 \\
                              & CLIP-AE & 45.3 & 28.7 & 6.7  & 28.9 \\
\hline
\end{tabular}
}
\end{table}

%% file: table/table03.tex
\begin{table}[h]
\centering
\caption{Albaltion studies on components of our framework}
\label{tab:my-table-3}
\resizebox{0.32\textwidth}{!}{
\begin{tabular}{c|c|ccc|c}
\hline
\multicolumn{1}{c|}{\multirow{2}{*}{CAF}} &
  \multicolumn{1}{c|}{\multirow{2}{*}{CCP}} &
  \multicolumn{3}{c|}{mAP@IoU (\%)} &
  \multicolumn{1}{c}
  {\multirow{2}{*}{CAF}}\\
\multicolumn{1}{c|}{} &
  \multicolumn{1}{c|}{} &
  \multicolumn{1}{c}{0.5} &
  \multicolumn{1}{c}{0.75} &
  \multicolumn{1}{c|}{0.95} &
  \multicolumn{1}{c}{} \\
\hline
\multicolumn{1}{c|}{} &
  \multicolumn{1}{c|}{} &
  \multicolumn{1}{c}{37.4} &
  \multicolumn{1}{c}{23.8} &
  \multicolumn{1}{c|}{4.9} &
  \multicolumn{1}{c}{22.7} \\
\multicolumn{1}{c|}{\checkmark} &
  \multicolumn{1}{c|}{} &
  \multicolumn{1}{c}{41.6} &
  \multicolumn{1}{c}{26.5} &
  \multicolumn{1}{c|}{5.6} &
  \multicolumn{1}{c}{25.1} \\
\multicolumn{1}{c|}{} &
  \multicolumn{1}{c|}{\checkmark} &
  \multicolumn{1}{c}{43.4} &
  \multicolumn{1}{c}{27.8} &
  \multicolumn{1}{c|}{6.1} &
  \multicolumn{1}{c}{26.5} \\
\multicolumn{1}{c|}{\checkmark} &
  \multicolumn{1}{c|}{\checkmark} &
  \multicolumn{1}{c}{45.3} &
  \multicolumn{1}{c}{28.7} &
  \multicolumn{1}{c|}{6.7} &
  \multicolumn{1}{c}{28.9} \\
  \hline
\end{tabular}
}
\end{table}

%% file: Template.bbl
\begin{thebibliography}{10}

\bibitem{zhai2020two}
Yuanhao Zhai, Le~Wang, Wei Tang, Qilin Zhang, Junsong Yuan, and Gang Hua,
\newblock ``Two-stream consensus network for weakly-supervised temporal action localization,''
\newblock in {\em Computer Vision--ECCV 2020: 16th European Conference, Glasgow, UK, August 23--28, 2020, Proceedings, Part VI 16}. Springer, 2020, pp. 37--54.

\bibitem{yun2024weakly}
Wulian Yun, Mengshi Qi, Chuanming Wang, and Huadong Ma,
\newblock ``Weakly-supervised temporal action localization by inferring salient snippet-feature,''
\newblock in {\em Proceedings of the AAAI Conference on Artificial Intelligence}, 2024, vol.~38, pp. 6908--6916.

\bibitem{wang2023two}
Yu~Wang, Yadong Li, and Hongbin Wang,
\newblock ``Two-stream networks for weakly-supervised temporal action localization with semantic-aware mechanisms,''
\newblock in {\em Proceedings of the IEEE/CVF Conference on Computer Vision and Pattern Recognition}, 2023, pp. 18878--18887.

\bibitem{zhang2024can}
Quan Zhang and Yuxin Qi,
\newblock ``Can mllms guide weakly-supervised temporal action localization tasks?,''
\newblock {\em arXiv preprint arXiv:2411.08466}, 2024.

\bibitem{Zhang2025Rethinking}
Quan Zhang, Yuxin Qi, Xi~Tang, Rui Yuan, Xi~Lin, Ke~Zhang, and Chun Yuan,
\newblock ``Rethinking pseudo-label guided learning for weakly supervised temporal action localization from the perspective of noise correction,''
\newblock {\em Proceedings of the AAAI Conference on Artificial Intelligence}, vol. 39, no. 10, pp. 10085--10093, Apr 2025.

\bibitem{liu2023apsl}
Yuanyuan Liu, Ning Zhou, Fayong Zhang, Wenbin Wang, Yu~Wang, Kejun Liu, and Ziyuan Liu,
\newblock ``Apsl: Action-positive separation learning for unsupervised temporal action localization,''
\newblock {\em Information Sciences}, vol. 630, pp. 206--221, 2023.

\bibitem{gong2020learning}
Guoqiang Gong, Xinghan Wang, Yadong Mu, and Qi~Tian,
\newblock ``Learning temporal co-attention models for unsupervised video action localization,''
\newblock in {\em Proceedings of the IEEE/CVF Conference on Computer Vision and Pattern Recognition}, 2020, pp. 9819--9828.

\bibitem{yang2022uncertainty}
Wenfei Yang, Tianzhu Zhang, Yongdong Zhang, and Feng Wu,
\newblock ``Uncertainty guided collaborative training for weakly supervised and unsupervised temporal action localization,''
\newblock {\em IEEE Transactions on Pattern Analysis and Machine Intelligence}, vol. 45, no. 4, pp. 5252--5267, 2022.

\bibitem{liu2023revisiting}
Qinying Liu, Zilei Wang, Shenghai Rong, Junjie Li, and Yixin Zhang,
\newblock ``Revisiting foreground and background separation in weakly-supervised temporal action localization: A clustering-based approach,''
\newblock in {\em Proceedings of the IEEE/CVF International Conference on Computer Vision}, 2023, pp. 10433--10443.

\bibitem{nguyen2018weakly}
Phuc Nguyen, Ting Liu, Gautam Prasad, and Bohyung Han,
\newblock ``Weakly supervised action localization by sparse temporal pooling network,''
\newblock in {\em Proceedings of the IEEE conference on computer vision and pattern recognition}, 2018, pp. 6752--6761.

\bibitem{nguyen2019weakly}
Phuc~Xuan Nguyen, Deva Ramanan, and Charless~C Fowlkes,
\newblock ``Weakly-supervised action localization with background modeling,''
\newblock in {\em Proceedings of the IEEE/CVF international conference on computer vision}, 2019, pp. 5502--5511.

\bibitem{ren2023proposal}
Huan Ren, Wenfei Yang, Tianzhu Zhang, and Yongdong Zhang,
\newblock ``Proposal-based multiple instance learning for weakly-supervised temporal action localization,''
\newblock in {\em Proceedings of the IEEE/CVF conference on computer vision and pattern recognition}, 2023, pp. 2394--2404.

\bibitem{zhang2025imdprompter}
Quan Zhang, Yuxin Qi, Xi~Tang, Jinwei Fang, Xi~Lin, Ke~Zhang, and Chun Yuan,
\newblock ``{IMDP}rompter: Adapting {SAM} to image manipulation detection by cross-view automated prompt learning,''
\newblock in {\em The Thirteenth International Conference on Learning Representations}, 2025.

\bibitem{chen2025gim}
Yirui Chen, Xudong Huang, Quan Zhang, Wei Li, Mingjian Zhu, Qiangyu Yan, Simiao Li, Hanting Chen, Hailin Hu, Jie Yang, et~al.,
\newblock ``Gim: A million-scale benchmark for generative image manipulation detection and localization,''
\newblock in {\em Proceedings of the AAAI Conference on Artificial Intelligence}, 2025, vol.~39, pp. 2311--2319.

\bibitem{qi2025seeing}
Yuxin Qi, Quan Zhang, Xi~Lin, Xiu Su, Jiani Zhu, Jingyu Wang, and Jianhua Li,
\newblock ``Seeing beyond noise: Joint graph structure evaluation and denoising for multimodal recommendation,''
\newblock in {\em Proceedings of the AAAI Conference on Artificial Intelligence}, 2025, vol.~39, pp. 12461--12469.

\bibitem{Zhang_2024_CVPR}
Quan Zhang, Xiaoyu Liu, Wei Li, Hanting Chen, Junchao Liu, Jie Hu, Zhiwei Xiong, Chun Yuan, and Yunhe Wang,
\newblock ``Distilling semantic priors from sam to efficient image restoration models,''
\newblock in {\em Proceedings of the IEEE/CVF Conference on Computer Vision and Pattern Recognition (CVPR)}, June 2024, pp. 25409--25419.

\end{thebibliography}
